\documentclass[sigconf]{acmart}

\AtBeginDocument{%
  }

\setcopyright{acmlicensed}
\copyrightyear{2026}
\acmYear{2026}
\acmDOI{10.1145/3805622.3810877}
\acmConference[ICMR '26]{the 2026 International Conference on Multimedia Retrieval}{June 16-19, 2026}{Amsterdam, Netherlands}
\acmISBN{978-1-4503-XXXX-X/2018/06}

\usepackage[nameinlink]{cleveref}
\usepackage{tabularx}
\usepackage{pifont}
\newcommand{\xmark}{\ding{55}}
\newcommand{\cmark}{\ding{51}}
\usepackage{multirow}
\usepackage[table]{xcolor}
\usepackage{makecell}
\usepackage{graphicx}
\usepackage{subcaption}

\begin{document}

\title{AIFIND: Artifact-Aware Interpreting Fine-Grained Alignment for Incremental Face Forgery Detection}

\author{Hao Wang, Beichen Zhang$^{*}$, Yanpei Gong, Shaoyi Fang, Zhaobo Qi, Yuanrong Xu, 
Xinyan Liu, Weigang Zhang}

\thanks{$^\ast$Corresponding author.}

\affiliation{
    \institution{School of Computer Science and Technology, Harbin Institute of Technology, Weihai}
    \country{}
}

\email{{2023210984, 2023211640, shaoyi.fang}@stu.hit.edu.cn}
\email{{beiczhang, qizb, xuyuanrong, xinyliu, wgzhang}@hit.edu.cn}









\begin{abstract}
As forgery types continue to emerge consistently, Incremental Face Forgery Detection (IFFD) has become a crucial paradigm. However, existing methods typically rely on data replay or coarse binary supervision, which fails to explicitly constrain the feature space, leading to severe feature drift and catastrophic forgetting. 
To address this, we propose AIFIND, Artifact-Aware Interpreting Fine-Grained Alignment for Incremental Face Forgery Detection, which leverages semantic anchors to stabilize incremental learning. We design the Artifact-Driven Semantic Prior Generator to instantiate invariant semantic anchors, establishing a fixed coordinate system from low-level artifact cues. These anchors are injected into the image encoder via Artifact-Probe Attention, which explicitly constrains volatile visual features to align with stable semantic anchors. Adaptive Decision Harmonizer harmonizes the classifiers by preserving angular relationships of semantic anchors, maintaining geometric consistency across tasks.
Extensive experiments on multiple incremental protocols validate the superiority of AIFIND.
\end{abstract}

\ccsdesc[500]{Security and privacy~Human and societal aspects of security and privacy}

\keywords{Face Forgery Detection, Incremental Learning, Semantic Anchors, Fine-Grained Visual-Text Alignment}

\maketitle

\section{Introduction}

With the rapid development of generative models, face forgery has advanced and achieved unprecedented realism, posing serious public safety risks. Thus, developing forgery detection becomes crucial for maintaining security in the digital ecosystem. Existing mainstream face forgery detection methods~\cite{lin2025standing, sun2025towards, forensics, zhou2024freqblender,shiohara2022detecting} attempt to train a generalizable detector with limited and static datasets. However, new forgery techniques emerge endlessly, which quickly render existing detection methods ineffective. In view of this, Incremental Face Forgery Detection (IFFD) is proposed to continuously train the model with the latest forged samples.

\begin{figure}[t]
    \centering
    \includegraphics[width=\linewidth]{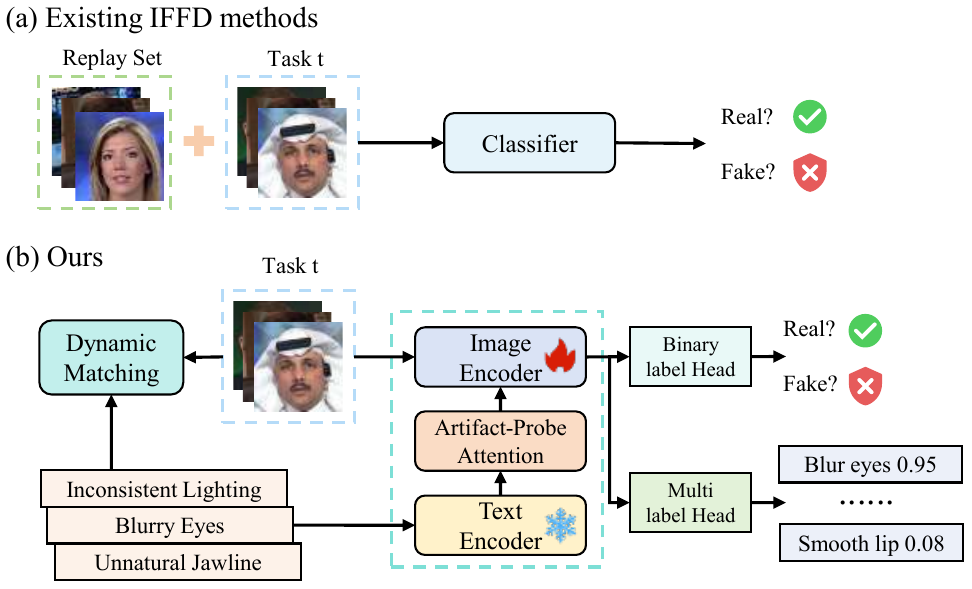}

    \caption{Comparison between AIFIND and other methods. (a) Conventional methods depend on replay sets to mitigate forgetting. (b) Our method realizes a data-replay-free paradigm via stable semantic anchors. Visual features are continuously aligned with stable semantic priors, ensuring consistent decision boundaries across tasks.}
    \label{fig:introduction}

\end{figure}

Currently, existing methods~\cite{pan2023dfil, tian2024dynamic, cheng2025stacking} for IFFD mainly rely on data replay, which preserves knowledge by storing a small subset of past samples. Representative approaches such as DFIL~\cite{pan2023dfil} and SUR-LID~\cite{cheng2025stacking} follow this paradigm. However, merely replaying discrete samples fails to explicitly constrain the topology of the feature space. Regardless of data replay or regularization, as shown in Fig.~\ref{fig:introduction}(a), current IFFD methods rely on coarse binary supervision, failing to leverage fine-grained artifact cues. Critically, in incremental learning settings, without stable anchors to stabilize the latent space, the learned feature distribution is prone to unconstrained drift. Consequently, when accommodating new forgery types, the model inadvertently overwrites previously learned representations, resulting in catastrophic forgetting.

Furthermore, while recent works~\cite{lin2025standing, forensics, sun2025towards} have sought to integrate Vision-Language Models (VLMs) to enhance deepfake detection, they primarily focus on static scenarios and fail to apply them in incremental learning. Existing VLM-based methods often treat semantic information as a coarse global label, neglecting the fine-grained semantic discrepancies between authentic and forged facial regions. 
Motivated by this, we rethink the IFFD paradigm by exploiting the intrinsic semantic sensitivity of pre-trained VLMs toward specific artifact-prone regions. 
We posit that linguistic concepts possess a natural invariance: while the visual manifestations of local anomalies vary across datasets, the semantic distinction between authentic and manipulated facial components remains constant. Leveraging this property, we propose to utilize these region-aware authenticity priors as invariant semantic anchors. By anchoring visual features to these stable semantic coordinates, we can explicitly stabilize the evolving visual feature space.

As shown in Fig.~\ref{fig:introduction}(b), we interpret artifacts as invariant semantic anchors. This formulation establishes a stable, high-dimensional reference for fine-grained supervision, enabling semantically guided artifact inference to resist feature drift.
Based on this, we propose \textbf{A}rtifact-Aware \textbf{I}nterpreting \textbf{F}ine-Grained Alignment for \textbf{In}cremental Face Forgery \textbf{D}etection (\textbf{AIFIND}). 
AIFIND consists of the following cooperative components:
(1) Artifact-Driven Semantic Prior Generator (\textbf{ASPG}) constructs the initial semantic anchors by interpreting volatile low-level artifact cues into stable sparse textual labels; 
(2) Artifact-Probe Attention (\textbf{APA}) module injects selected textual artifact cues into the image encoder for fine-grained visual–text alignment;
(3) Semantic-Guided Incremental Detector (\textbf{SGID}) governs the learning process within this anchored space, leveraging APA and dual supervision to simultaneously discriminate authenticity and identify specific artifact types; and 
(4) Adaptive Decision Harmonizer (\textbf{ADH}) aligns binary and multi-label heads by strictly preserving the angular relationships relative to the semantic anchors, guaranteeing consistency in decision boundaries.

During training, AIFIND executes a dynamic matching strategy to achieve stable incremental learning.
First, ASPG instantiates semantic anchors to form a fixed coordinate system. The model then transitions to a dynamic matching mechanism, autonomously recalibrating targets based on similarity. 
Then, APA injects these matched semantic anchors to enforce fine-grained anchoring. This process continuously rectifies volatile visual features via stable semantic definitions. 
Finally, ADH harmonizes the classifiers by preserving their angular relationships relative to the semantic anchors, maintaining geometric consistency across tasks.
As a result, AIFIND effectively mitigates catastrophic forgetting, enabling semantically coherent knowledge evolution without replay buffers.

Experiments under multiple incremental protocols~\cite{cheng2025stacking} validate the superiority and strong generalization capability of our method. 
Our main contributions are summarized as follows:
\begin{itemize}
    \item We propose a data-replay-free framework for IFFD, which reinterprets artifacts as semantic anchors to explicitly constrain the feature space without storing past data.
    
    \item We formulate an artifact-probe attention that effectively anchors evolving visual forgery cues to immutable semantic anchors, preventing feature drift.
    
    \item We conduct comprehensive experiments that empirically validate the superiority of our framework, proving the effectiveness of semantic anchors in incremental learning.

\end{itemize}

\section{Related Work}

\subsection{Face Forgery Detection}
Face forgery detection has become a prominent research topic in computer vision. 
Early studies mainly relied on detecting anomalies in biometric cues such as eye blinking~\cite{ictu}, head pose~\cite{yang2019exposing}, and pupil morphology~\cite{guo2022eyes}. 
With the rapid advancement of deep learning, attention gradually shifted toward learning forgery traces from multimodal clues, such as frequency domain~\cite{zhou2024freqblender,kashiani2025freqdebias} and temporal consistency~\cite{yan2025generalizing,guo2025face}. 
Recently, the emergence of vision–language models such as CLIP~\cite{clip} has opened up new possibilities for semantic-level deepfake detection. 
RepDFD~\cite{lin2025standing} reprograms CLIP by injecting universal perturbations into input images, while ForAda~\cite{forensics} leverages a Forensics Adapter to learn hybrid boundary artifacts specific to facial forgeries. VLFFD~\cite{sun2025towards} attempts fine-grained visual–text alignment by incorporating detailed textual semantics, but its dependence on paired real–fake data limits its applicability to incremental learning. These approaches~\cite{zhou2024freqblender,shiohara2022detecting,yan2025generalizing,yan2024transcending,forensics} aim to extract universal forgery representations from limited observed data and achieve effective transfer to unseen manipulation types.

\subsection{Incremental Face Forgery Detection} 

As forgery techniques evolve rapidly, constructing a general detector from limited training datasets has become increasingly impractical. 
This motivates the exploration of Incremental Face Forgery Detection (IFFD), which enables models to continuously learn emerging forgery patterns while preserving previous knowledge.

Incremental learning methods are commonly categorized into three paradigms: parameter isolation~\cite{wang2023rehearsal,wang2025boosting}, parameter regularization~\cite{sarkar2025adaptive,batra2024evcl}, and data replay~\cite{buzzega2021rethinking,smith2024adaptive,mandalika2025replay}. 
In the context of IFFD, most existing studies emphasize knowledge distillation and replay mechanisms. 
CoReD~\cite{kim2021cored} preserves old knowledge through task-aware distillation while adapting to new domains. 
DFIL~\cite{pan2023dfil} replays representative and challenging samples from previous datasets, 
and DMP~\cite{tian2024dynamic} dynamically expands prototypes to accommodate newly emerging forgery types. 
HDP~\cite{sun2025continual} achieves replay through UAPs~\cite{moosavi2017universal}, 
while SUR-LID~\cite{cheng2025stacking} aligns latent features across tasks to mitigate mutual interference. 
However, these methods still treat all forgery instances as a single \textit{Fake} category, which limits their ability to capture fine-grained artifact semantics.

Meanwhile, Vision–Language Models (VLMs) such as CLIP~\cite{unlocking} have inspired new paradigms for incremental learning. 
Prompt-based approaches~\cite{roy2024convolutional,wang2022learning,kim2024one,coda} design task-specific prompts to guide CLIP models, 
while adapter-based techniques~\cite{li2025dynamic,yu2024boosting,lora} insert lightweight trainable modules at intermediate layers to encapsulate new knowledge. 
Despite their success in general incremental learning, these methods are not specifically designed for the unique challenges of deepfake detection, where subtle visual cues play a critical role, leading to limited effectiveness when applied directly.

\begin{figure*}[!t]
    \centering
    \includegraphics[width=\textwidth]{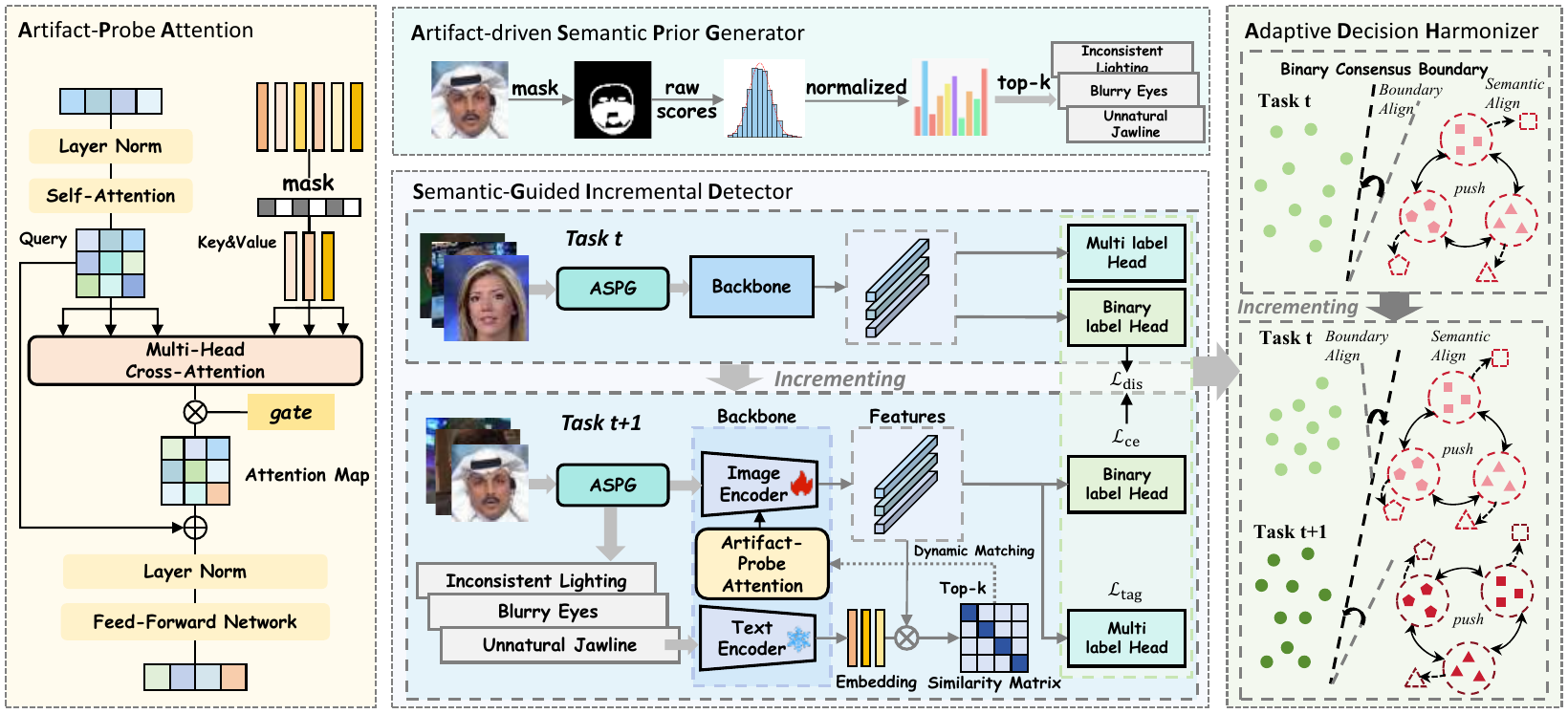}
    \caption{Overall framework of AIFIND. Artifact-Driven Semantic Prior Generator (ASPG) instantiates semantic anchors to build a fixed coordinate system. Semantic-Guided Incremental Detector (SGID) uses the Artifact-Probe Attention (APA) to perform the anchoring, which constrains volatile visual features to these stable semantic anchors. Adaptive Decision Harmonizer (ADH) maintains geometric semantic consistency, preserving semantic angular relationships across tasks.}
    \label{fig:overview}
\end{figure*}

\section{Method}
\subsection{Framework overview}

As illustrated in Fig.~\ref{fig:overview}, we introduce AIFIND to enable stable incremental learning via a semantically anchored feature space. Specifically, ASPG transforms low-level cues into semantic anchors. These anchors are injected into image encoder by SGID using APA to ensure fine-grained visual–text alignment. Finally, ADH aligns classifier weights to preserve geometric consistency across tasks.

\subsection{Artifact-Driven Semantic Prior Generator}
According to ~\cite{sun2025towards} and our analysis, we define 5 representative forgery dimensions 
$\mathcal{I}=\{\mathcal{I}_{\text{blur}}, \mathcal{I}_{\text{color}}, \mathcal{I}_{\text{structure}}, \allowbreak \mathcal{I}_{\text{text}}, \mathcal{I}_{\text{boundary}}\}$. 
To spatially ground these dimensions, we utilize \textit{MediaPipe Face Mesh} \allowbreak 
 \cite{mediapipe} to locate 6 facial regions 
$\mathcal{R}=\{R_{\text{eyes}}, R_{\text{nose}}, R_{\text{cheeks}}, \allowbreak R_{\text{mouth}}, R_{\text{jawline}}, \allowbreak  R_{\text{boundary}}\}$. 
Additionally, a global \textit{skin reference region} $S$ is extracted to serve as the baseline for computing relative inconsistencies.
The correspondence between facial regions and indicators is in Tab.~\ref{tab:indicator} and definitions of indicators are as follows:

\begin{table}[t]
    \centering
    \caption{Correspondence between Indicators and Facial Regions. \cmark~indicates that the specific operator is applied.}
    \label{tab:indicator}
    \footnotesize

    \renewcommand{\arraystretch}{1} 
    \newcolumntype{Y}{>{\centering\arraybackslash}X} 
    
    \begin{tabularx}{\linewidth}{c Y Y Y Y Y}
        \toprule
        Facial Region & Blur & Color & Structure & Texture & Boundary \\
        \midrule
        Eyes     & \cmark & \cmark & \cmark & \cmark & \xmark \\ 
        Nose     & \cmark & \cmark & \cmark & \cmark & \xmark \\ 
        Cheeks   & \cmark & \cmark & \cmark & \cmark & \xmark \\
        Mouth    & \cmark & \cmark & \cmark & \cmark & \xmark \\ 
        Jawline  & \xmark & \xmark & \xmark & \xmark & \cmark \\
        Boundary & \xmark & \xmark & \xmark & \xmark & \cmark \\ 
        \bottomrule
    \end{tabularx}
\end{table}

\textbf{Blur Indicator.} 
We measure local sharpness using the variance of the Laplacian operator:
\begin{equation}
    \mathcal{I}_{\text{blur}} = \mathrm{Var}\!\left(\nabla^2 I_{\mathcal{M}}\right),
\end{equation}
where $I_{\mathcal{M}}$ denotes the grayscale intensities within the region mask.

\textbf{Color Indicator.} 
To capture lighting inconsistencies, we calculate the luminance deviation between the target region $R$ and the skin reference $S$ in CIELAB space:
\begin{equation}
    \mathcal{I}_{\text{color}} = \left| L_{R} - L_{S} \right|,
\end{equation}
where $L_R$ and $L_S$ represent the mean $L$-channel values, respectively.

\textbf{Structural Indicator.} 
We assess the structural compatibility between the target region and the surrounding skin using the Structural Similarity Index (SSIM):
\begin{equation}
    \mathcal{I}_{\text{struct}} = \mathrm{SSIM}\!\left(P_R, P_S\right),
\end{equation}
where $P_R$ and $P_S$ are normalized grayscale patches from $R$ and $S$.

\textbf{Texture Indicator.} 
To expose statistical anomalies in skin texture, we extract local contrast using the Gray-Level Co-occurrence Matrix (GLCM):
\begin{equation}
    \mathcal{I}_{\text{text}} = \mathrm{Contrast}\!\left(\mathrm{GLCM}\!\left(I_{\mathcal{M}}\right)\right),
\end{equation}
where $I_{\mathcal{M}}$ is quantized to 64 levels to retain details.

\textbf{Boundary Indicator.} 
We identify potential blending artifacts along edges by computing the average gradient magnitude:
\begin{equation}
    \mathcal{I}_{\text{bound}} = \mathbb{E}_{(x,y)\in \mathcal{M}} \left[\sqrt{G_x^2 + G_y^2}\right],
\end{equation}
where $G_x$ and $G_y$ denote Sobel derivatives.

To bridge pixel-level statistics with high-level semantics, for each artifact dimension $i \in \mathcal{I}$ and region $g \in \mathcal{R}$, we leverage an LLM to generate a candidate set of $K$ contrastive text pairs, denoted as $\mathcal{T}_{i, g} = \{ (t^{\text{r}}_k, t^{\text{f}}_k) \}_{k=1}^K$, and align them with a support set $\Omega_{i,g}$. The optimal anchor $A_{i,g}$ is selected by calculating CLIP similarity:
\begin{equation}
    A_{i,r} = \operatorname*{argmax}_{(t^{\text{r}}, t^{\text{f}}) \in \mathcal{T}_{i,g}} \sum_{(x^{\text{r}}, x^{\text{f}}) \in \Omega_{i,g}} \left[ \operatorname{Sim}(t^{\text{f}}, x^{\text{f}}) + \operatorname{Sim}(t^{\text{r}}, x^{\text{r}}) \right],
\end{equation}
where $r$ refers to \textit{real} and $f$ refers to \textit{fake}.
By iterating this process across all indicators and facial regions, we construct a  \textit{Semantic Anchor Library} $\mathcal{A} = \{ A_{i,g} \mid i \in \mathcal{I}, g \in \mathcal{R} \}$.

For a forged image $x^{\text{f}}$, we target the most severe anomalies by selecting the top-$N$ highest scores, and retrieve their corresponding forgery descriptions $t^{\text{f}}$. 
Conversely, for a real image $x^{\text{r}}$, we focus on the most pristine facial details by selecting the bottom-$N$ lowest scores, assigning the linked authentic descriptions $t^{\text{r}}$.

\subsection{Semantic-Guided Incremental Detector}
Building upon the instantiated semantic anchors, we propose the Semantic-Guided Incremental Detector (SGID) to explicitly inject these semantic anchors into the visual learning process. The framework is underpinned by two components: \textbf{Artifact-Probe Attention (APA)} and \textbf{Dual Supervision}.

\textbf{Artifact-Probe Attention.}
To effectively incorporate the semantic priors, we introduce the APA module within the vision transformer.
Let $X \in \mathbb{R}^{P \times D}$ denote the intermediate visual embeddings, where $P$ is the number of patches.
Simultaneously, let $S = \Phi_{\text{text}}(\mathcal{A}_{matched}) \in \mathbb{R}^{N \times D}$ represent the textual embeddings of the matched semantic anchors, where $N$ is the number of selected anchors.
The APA module functions as a cross-modal bridge, employing a Multi-Head Attention mechanism where visual tokens serve as queries to probe the semantic details in the embeddings:
\begin{equation}
\tilde{X} = \mathrm{MHA}(Q = X, K = S, V = S).
\label{eq:mhca}
\end{equation}

To ensure adaptive integration, we employ a residual connection with a learnable gating parameter $g \in \mathbb{R}^D$:
\begin{equation}
X_{\text{fused}} = X + g \odot \tilde{X},
\label{eq:apa_fusion}
\end{equation}
where $\odot$ denotes element-wise multiplication. The gating coefficient $g$ dynamically modulates the injection of semantic priors, preventing the overriding of intrinsic visual cues. The fused representation $X_{\text{fused}}$ then proceeds to the subsequent layers. In practice, APA is injected into the top-$M$ transformer layers.

\textbf{Dual Supervision.}
To enforce the alignment between visual features and the selected semantic anchors, we employ a dual supervision strategy.
Given an input image $x$, let $F = \Phi_{\text{img}}(x, S)$ denote the visual features extracted by the APA-enhanced image encoder.

First, a binary classifier $C_t$ is employed to predict global authenticity, formulated as:
\begin{equation}
    \mathcal{L}_{\text{cls}} = \mathrm{CE}(C_t(F), Y_{\text{bin}}),
    \label{eq:binary}
\end{equation}
where $Y_{\text{bin}} \in \{0,1\}$ is the ground-truth label (Real/Fake), and $\mathrm{CE}(\cdot)$ denotes the standard cross-entropy loss.

Simultaneously, to ensure the model comprehends the specific forgery patterns described by the semantic anchors, a multi-label head $H_t$ is utilized to predict the presence of the defined artifact dimensions ($\mathcal{I}_{\text{blur}}, \mathcal{I}_{\text{color}}, \mathcal{I}_{\text{structure}}, \mathcal{I}_{\text{text}}, \mathcal{I}_{\text{boundary}}$). 
The artifact dimensions prediction loss is defined as:
\begin{equation}
    \mathcal{L}_{\text{ind}} = \mathrm{BCE}(H_t(F), Y_{\text{ind}}),
    \label{eq:tag}
\end{equation}
where $Y_{\text{ind}} \in \{0,1\}^{|\mathcal{I}|}$ denotes the binary artifact indicator vector corresponding to the dimensions in $\mathcal{I}$, and $\mathrm{BCE}(\cdot)$ is the binary cross-entropy loss applied independently to each attribute. 

By jointly optimizing $\mathcal{L}_{\text{cls}}$ and $\mathcal{L}_{\text{ind}}$, the model is encouraged to align its visual features with the stable semantic anchors. 
The stationary semantic space acts as a persistent reference throughout the incremental learning process, effectively preventing feature drift and mitigating catastrophic forgetting.

\subsection{Adaptive Decision Harmonizer}

In incremental learning, the decision boundaries of classifiers tend to drift as new tasks are introduced. To mitigate this, we propose ADH to align classifier weights to preserve geometric consistency across tasks.

Let $\tilde{W}^{(b)}_j$ and $\tilde{W}^{(m)}_j$ respectively denote the normalized classifier weights of the binary-label and multi-label heads from previous tasks. 
To quantify the semantic affinity between the current task and historical knowledge, ADH computes adaptive similarity weights for each head $h \in \{b, m\}$:
\begin{equation}
    \omega_j^{(h)} = \frac{\exp\!\big(\cos(\tilde{W}^{(h)}_i,\tilde{W}^{(h)}_j)/\tau\big)}
    {\sum_{l \neq i}\exp\!\big(\cos(\tilde{W}^{(h)}_i,\tilde{W}^{(h)}_l)/\tau\big)},
    \label{eq:omega_softmax}
\end{equation}
where $\tau$ is a temperature parameter and $j \neq i$. High similarity implies that the current artifacts share underlying semantic traits with task $j$, warranting stronger alignment.

We then construct a \textit{Global Semantic Reference} $\tilde{W}_{\text{ref}}^{(h)}$ by aggregating previous classifiers according to their semantic affinity:
\begin{equation}
    \tilde{W}_{\text{ref}}^{(h)} =
    \mathrm{norm}\!\left(
    \sum_{j \neq i}
    \omega_j^{(h)} \tilde{W}^{(h)}_j
    \right).
    \label{eq:ref_head}
\end{equation}
This reference vector captures the stable historical decision trend, serving as a robust alignment anchor to prevent the model from forgetting previously learned forgery patterns.

To incorporate the historical knowledge without distorting the feature space, we perform a \textit{spherical semantic alignment}. Unlike Euclidean interpolation, this process respects the geometric structure of the hypersphere, rotating the current decision boundary toward the global reference along the geodesic path:
\begin{equation}
    \tilde{W}_{\text{new}}^{(h)} =
    \frac{\sin((1-t^{(h)})\theta)}{\sin\theta}\tilde{W}_i^{(h)} +
    \frac{\sin(t^{(h)}\theta)}{\sin\theta}\tilde{W}_{\text{ref}}^{(h)},
    \label{eq:slerp_merge}
\end{equation}
where $\theta$ represents the angular distance between the current and reference weights. 
Crucially, the alignment coefficient $t^{(h)} \in [0,1]$ is adaptively determined to balance plasticity and stability:
\begin{equation}
    t^{(h)} =\cos(\tilde{W}_i^{(h)}, \tilde{W}_{\text{ref}}^{(h)}),
    \label{eq:t}
\end{equation}
 where a high cosine similarity indicates that the current task is semantically consistent with history, triggering a gentle update to preserve existing anchors.

Finally, to decouple the semantic directional alignment from the magnitude-dependent feature strength, we rescale the aligned weights to recover the norm of $W_i^{(h)}$:
\begin{equation}
    W_{\text{new}}^{(h)} =
    \tilde{W}_{\text{new}}^{(h)} \cdot \| W_i^{(h)} \|_2,
    \label{eq:rescale}
\end{equation}
which ensures that the alignment modifies only the semantic direction without degrading the detector's discriminative sensitivity. 

By constraining the update trajectory to the spherical manifold, ADH prevents the decision boundaries from drifting into semantically ambiguous regions, guaranting that the evolving detector remains geometrically consistent and semantically coherent with the accumulated global anchor library.

\subsection{Training strategy and overall loss}
\textbf{Training strategy.}
During the initial $n$ epochs, the semantic anchors $S$ are \textbf{fixed} to the  descriptions generated by ASPG, providing stable and consistent guidance to the backbone. 
Starting from the $(n+1)$-th epoch, we activate the \textbf{dynamic matching} mechanism. In this phase, $S$ is adaptively retrieved by calculating the cosine similarity between visual features and textual embeddings, enabling the model to capture fine-grained, instance-specific forgery patterns.

\noindent \textbf{Overall loss.}
Following ~\cite{pan2023dfil}, we also maintain the previous-task learned information via knowledge distillation loss, which is:
\begin{equation}
    \mathcal{L}_{\text{dis}} = \| \Phi_{\text{img}}(x, S; \theta_t) - \Phi_{\text{img}}(x, S; \theta_{t-1}) \|_2^2,
    \label{eq:dist}
\end{equation}
where $\Phi_{\text{img}}(\cdot; \theta_{t-1})$ is the frozen backbone extractor trained on the previous $(t-1)$-th task, which serves as a reference to regularize the feature space on the current data.

The total loss function is defined as:
\begin{equation}
\mathcal{L}_{\text{overall}} = 
\mathcal{L}_{\text{cls}} + \mu_1 \mathcal{L}_{\text{ind}} + \mu_2 \mathcal{L}_{\text{dis}},
\label{eq:total_loss}
\end{equation}
where $\mu_1$ and $\mu_2$ are trade-off coefficients.

\section{Experiment}
\subsection{Experimental Settings}
\textbf{Datasets.} Our datasets follow the protocol in ~\cite{cheng2025stacking} to ensure a comprehensive evaluation. We use several forgery face datasets, including DeepFake Detection Challenge Preview (DFDCP)~\cite{dfdcp}, Celeb-DF-v2 (CDF)~\cite{celeb}, and FaceForensics++~\cite{ff++}. Furthermore, we incorporate recently released datasets that feature more diverse and sophisticated forgery methods, namely {MCNet~\cite{mcnet}, BlendFace~\cite{blendface}, StyleGAN3~\cite{gan}} from DF40~\cite{df40} and {SDv21~\cite{sdv21}} from DiffusionFace~\cite{diffusionface}.

\noindent\textbf{Evaluation Protocol.} 
To systematically evaluate the performance of our model, we adopt the standard evaluation protocols in~\cite{cheng2025stacking}. 

\begin{itemize}
    \item \textbf{Protocol 1 (P1): Datasets Incremental} with \{SDv21, FF++, DFDCP, CDF\}. This protocol simulates scenarios in which the model is required to adapt to entirely novel data environments across different incremental steps. 
    \item \textbf{Protocol 2 (P2): Forgery Categories Incremental} with \{Hybrid (FF++), Face-Reenactment (MCNet), Face-Swapping (BlendFace), Entire Face Synthesis (StyleGAN3)\}. The model continuously learns to defend against new forgery types, while the distribution of genuine data remains constant. 
\end{itemize}

\noindent\textbf{Implementation Details.} We use CLIP-ViT-L/14~\cite{clip} as backbone, fine-tuned via LN-tuning~\cite{unlocking}. The Adam optimizer is employed with a learning rate of $8 \times 10^{-5}$, 20 epochs, and batch size of 32. For replay-based baselines, the replay buffer size is set to 500 for each task. We set $\mu_1\!=\!0.1$, $\mu_2\!=\!1$ and set the number of selected semantic anchors to $N=3$. To ensure training stability, we set the warm-up period to $n=5$. The APA modules are injected into the last $M=4$ layers of the image encoder.
We employ Frame-level Area Under the Curve (AUC) as the evaluation metric. 

\begin{table*}[]
\centering
\footnotesize
\caption{
Performance comparisons (AUC) with Protocol 1 (Datasets Incremental) and Protocol 2 (Forgery Categories Incremental). 
Task 1 (T1) to Task 4 (T4) represent current incremented tasks in \{SDv21, FF++, DFDCP, CDF\} or \{Hybrid, FR, FS, EFS\}. 
The \textbf{bold} denotes the best ones. 
\textdagger \ \ indicates results copied from ~\cite{cheng2025stacking}.
\ddag \ \ indicates that the backbones of DFIL~\cite{pan2023dfil} and SUR-LID~\cite{cheng2025stacking} are replaced with ViT-L/14 in the experiments for a fair comparison.
}
\vspace{-5pt}
\begin{tabular*}{\textwidth}{@{\extracolsep{\fill}}l@{\hspace{3pt}}cc|ccccc|ccccc}

\toprule
\multirow{2}{*}{Method} & \multirow{2}{*}{Replays} & \multirow{2}{*}{Task}  
& \multicolumn{5}{c|}{Protocol 1} & \multicolumn{5}{c}{Protocol 2} \\ 
\cmidrule(lr){4-8} \cmidrule(lr){9-13}
& & & SDv21 & FF++ & DFDCP & CDF & Avg. & Hybrid & FR & FS & EFS & Avg. \\ 
\toprule

\rowcolor{gray!20}
\multicolumn{13}{c}{\textbf{Methods with CNN Backbone}} \\
\midrule

\multirow{4}{*}{CoReD (MM'21)\textsuperscript{\dag}~\cite{kim2021cored}} & \multirow{4}{*}{500} & T1 
& 0.9998 & - & - & - & 0.9998 & 0.9665 & - & - & - & 0.9665 \\ 
& & T2 
& 0.7459 & 0.9433 & - & - & 0.8446 & 0.9355 & 0.7988 & - & - & 0.8671 \\ 
& & T3 
& 0.8555 & 0.9096 & 0.8154 & - & 0.8602 & 0.8907 & 0.7929 & 0.8605 & - & 0.8480 \\ 
& & T4 
& 0.8718 & 0.8376 & 0.7987 & 0.9341 & 0.8606 & 0.8454 & 0.6429 & 0.8417 & 0.9263 & 0.8141 \\ 
\midrule

\multirow{4}{*}{DFIL (MM'23)\textsuperscript{\dag}~\cite{pan2023dfil}} & \multirow{4}{*}{500} & T1 
& 0.9998 & - & - & - & 0.9998 & 0.9646 & - & - & - & 0.9646 \\ 
& & T2 
& 0.7400 & 0.9466 & - & - & 0.8433 & 0.5574 & 0.9975 & - & - & 0.7775 \\ 
& & T3 
& 0.9692 & 0.8164 & 0.9088 & - & 0.8981 & 0.6071 & 0.6649 & 0.9903 & - & 0.7541 \\ 
& & T4 
& 0.9326 & 0.7397 & 0.7908 & 0.9881 & 0.8628 & 0.5083 & 0.9556 & 0.7081 & 0.9996 & 0.7929 \\ 
\midrule

\multirow{4}{*}{HDP (IJCV'24)\textsuperscript{\dag}~\cite{sun2025continual}} & \multirow{4}{*}{500} & T1 
& 0.9998 & - & - & - & 0.9998 & 0.9671 & - & - & - & 0.9671 \\ 
& & T2 
& 0.8373 & 0.9507 & - & - & 0.8940 & 0.6741 & 0.9545 & - & - & 0.8143 \\ 
& & T3 
& 0.9341 & 0.8532 & 0.8737 & - & 0.8870 & 0.6300 & 0.7135 & 0.9509 & - & 0.7648 \\ 
& & T4 
& 0.9055 & 0.8039 & 0.8412 & 0.9501 & 0.8752 & 0.5989 & 0.7006 & 0.8934 & 0.9373 & 0.7826 \\ 
\midrule

\multirow{4}{*}{SUR-LID (CVPR'25)\textsuperscript{\dag}~\cite{cheng2025stacking}} & \multirow{4}{*}{500} & T1 
& 0.9999 & - & - & - & 0.9999 & 0.9685 & - & - & - & 0.9685 \\ 
& & T2 
& 0.9937 & 0.9485 & - & - & 0.9711 & 0.8291 & 0.9242 & - & - & 0.8766 \\ 
& & T3 
& 0.9986 & 0.8844 & 0.9161 & - & 0.9330 & 0.9050 & 0.9626 & 0.9794 & - & 0.9490 \\ 
& & T4 
& 0.9971 & 0.8479 & 0.9067 & 0.9744 & 0.9315 & 0.8790 & 0.9679 & 0.9356 & 0.9907 & 0.9433 \\ 
\midrule

\rowcolor{gray!20}
\multicolumn{13}{c}{\textbf{Methods with Vision Transformer}} \\
\midrule

\multirow{4}{*}{DFIL (ViT-L/14)\textsuperscript{\ddag}} & \multirow{4}{*}{500} & T1 
& 0.9999 & - & - & - & 0.9999 & 0.9682 & - & - & - & 0.9682 \\ 
& & T2 
& 0.8342 & 0.9581 & - & - & 0.8962 & 0.5831 & 0.9981 & - & - & 0.7906 \\ 
& & T3 
& 0.9897 & 0.8372 & 0.9192 & - & 0.9154 & 0.6598 & 0.7324 & 0.9941 & - & 0.7954 \\ 
& & T4 
& 0.9413 & 0.7523 & 0.8241 & 0.9923 & 0.8775 & 0.5748 & 0.9627 & 0.7597 & 0.9997 & 0.8242 \\ 
\midrule

\multirow{4}{*}{SUR-LID (ViT-L/14)\textsuperscript{\ddag}} & \multirow{4}{*}{500} & T1 
& 0.9999 & - & - & - & 0.9999 & 0.9406 & - & - & - & 0.9406 \\ 
& & T2 
& 0.9999 & 0.9090 & - & - & 0.9545 & 0.9235 & 0.9902 & - & - & 0.9568 \\ 
& & T3 
& 0.9997 & 0.9012 & 0.9238 & - & 0.9416 & 0.9043 & 0.9886 & 0.9756 & - & 0.9562 \\ 
& & T4 
& \textbf{0.9997} & 0.8838 & 0.9394 & 0.9714 & 0.9486 & 0.9035 & \textbf{0.9898} & 0.9773 & 0.9994 & 0.9675 \\ 
\midrule

\rowcolor{gray!20}
\multicolumn{13}{c}{\textbf{Traditional replay-free ViT-based Incremental Learning Methods}} \\
\midrule

\multirow{4}{*}{Coda-Prompt (CVPR'23)~\cite{coda}} & \multirow{4}{*}{0} & T1 
& 0.9999 & - & - & - & 0.9999 & 0.9631 & - & - & - & 0.9631 \\ 
& & T2 
& 0.7958 & 0.9323 & - & - & 0.8645 & 0.6132 & 0.8231 & - & - & 0.7181 \\ 
& & T3 
& 0.8564 & 0.8315 & 0.9026 & - & 0.8635 & 0.5816 & 0.7346 & 0.8831 & - & 0.7331 \\ 
& & T4 
& 0.8049 & 0.7492 & 0.8131 & 0.9413 & 0.8271 & 0.5245 & 0.5938 & 0.6264 & 0.9461 & 0.6727 \\ 
\midrule

\multirow{4}{*}{CL-LoRA (CVPR'25)~\cite{lora}} & \multirow{4}{*}{0} & T1 
& 0.9997 & - & - & - & 0.9997 & 0.9531 & - & - & - & 0.9531 \\ 
& & T2 
& 0.7892 & 0.9421 & - & - & 0.8656 & 0.6216 & 0.9846 & - & - & 0.8031 \\ 
& & T3 
& 0.8015 & 0.8896 & 0.8991 & - & 0.8634 & 0.5831 & 0.7216 & 0.9733 & - & 0.7593 \\ 
& & T4 
& 0.7742 & 0.8256 & 0.8549 & 0.9626 & 0.8543 & 0.5367 & 0.5966 & 0.7591 & 0.9988 & 0.7228 \\ 
\midrule

\rowcolor{gray!20}
\multicolumn{13}{c}{\textbf{Our Method}} \\
\midrule

\multirow{4}{*}{\textit{\textbf{AIFIND(Ours)}}} & \multirow{4}{*}{0} & T1 
& 0.9999 & - & - & - & 0.9999 & 0.9719 & - & - & - & 0.9719 \\ 
& & T2 
& 0.9999 & 0.9484 & - & - & 0.9742 & 0.9605 & 0.9948 & - & - & 0.9776 \\ 
& & T3 
& 0.9999 & 0.9397 & 0.9267 & - & 0.9554 & 0.9447 & 0.9909& 0.9817 & - & 0.9724 \\ 
& & T4 
& 0.9988 & \textbf{0.9327} & \textbf{0.9551} & \textbf{0.9879} & \textbf{0.9686} & \textbf{0.9286} & 0.9897 & \textbf{0.9892} & \textbf{0.9999} & \textbf{0.9769} \\

\bottomrule
\end{tabular*}
\label{tab:result}
\end{table*}

\subsection{Comparison with Other Methods}
To comprehensively evaluate the effectiveness of our proposed framework, we conduct extensive comparisons under Protocol~1 (Cross-Dataset) and Protocol~2 (Cross-Manipulation). 
As reported in Tab.\ref{tab:result}, our method consistently outperforms all baselines across both protocols, achieving a superior trade-off between retaining past knowledge and adapting to new forgeries.

\textbf{Comparison with Replay-based IFFD Methods.} 
Unlike replay-based approaches such as DFIL~\cite{pan2023dfil} and SUR-LID~\cite{cheng2025stacking}, which rely on storing historical samples to mitigate forgetting, our framework achieves better performance in a strict data-replay-free setting.
This effectively highlights the efficiency of semantic anchors. By substituting raw data storage with invariant semantic priors, we effectively circumvent the storage constraints and privacy concerns inherent in replay-based paradigms.

\textbf{Comparison with General Replay-free ViT-based Methods.} 
Directly transferring general Incremental Learning (IL) methods, including prompt-based (Coda-Prompt~\cite{coda}) and adapter-based (CL-LoRA~\cite{lora}) techniques, to the IFFD task leads to significant degradation. 
This reveals their inherent limitation: designed for object recognition, they primarily focus on high-level semantic content rather than subtle, low-level forgery \textit{artifacts}. 
Consequently, they fail to decouple forensic traces from content, lacking the fine-grained discriminative cues required for this specific task.

\textbf{Fairness Validation.} 
To ensure a rigorous comparison and rule out the influence of backbone capacity, we replace the backbones of DFIL~\cite{pan2023dfil} and SUR-LID~\cite{cheng2025stacking} with the ViT-L/14 used in our method. 
Even under this aligned configuration, our method still demonstrates clear superiority in performance. 
These results conclusively verify that our performance gains stem from the proposed methods rather than the raw power of the backbone. 
It underscores that the core challenge of IFFD lies in effective feature alignment, which cannot be solved solely by scaling up model parameters.

\subsection{Ablation Study}

\noindent \textbf{Overall Ablation.} 
As reported in Tab.\ref{tab:ablation}, the ablation study validates the indispensability of each component.
The absence of \textbf{ADH} results in significant decision boundary drift.
Removing \textbf{$\mathcal{L}_{\text{ind}}$} impairs feature disentanglement, degrading the model's ability to explicitly encode specific artifacts.
Furthermore, omitting \textbf{APA} weakens the cross-modal alignment, depriving the visual encoder of fine-grained semantic guidance.
These results demonstrate that optimizing both decision harmonization and semantic-visual interaction is essential for IFFD, enabling the unified framework to mitigate catastrophic forgetting while adapting to novel forgery patterns.

\noindent \textbf{Effect of Alignment Method.} 
We evaluate the impact of different classification-head alignment methods within ADH by comparing Linear (LERP), Exponential Moving Average (EMA), and Weighted Mean (WM) against our method. 
Unlike Euclidean-based methods that distort weight magnitudes, as shown in Tab.\ref{tab:adh}, our method yields superior performance, outperforming baselines that suffer from high-dimensional directional drift (LERP) or delayed adaptation (EMA). 
This confirms that strictly constraining the update trajectory to the spherical space preserves semantic angular consistency, ensuring geometrically coherent decision boundaries that robustly mitigate catastrophic forgetting across incremental tasks.

\noindent \textbf{Effect of APA Injection Layers.} 
We investigate the impact of APA injection depth by evaluating five configurations: Low-level (1–4), Medium-level (11–14), High (21–24), Multi-level (5, 10, 15, 20) and All Layers (0–24). 
As shown in Tab.~\ref{tab:apa}, the High-level setting yields superior performance, outperforming other configurations. 
This validates that semantic anchors align most effectively with high-level visual features that share similar semantic granularity, whereas other settings introduce semantic noise that tends to interfere with the extraction of high-level visual features.

\noindent \textbf{Hyperparameter Analysis.} 
We further investigate the impact of key hyperparameters: the number of selected semantic anchors $N$, the warm-up period $n$, the number of APA injection layers $M$ and the trade-off coefficients $\mu_1$ and $\mu_2$. 
As shown in Tab.\ref{tab:params}, $N=3$ achieves the optimal trade-off between guidance and noise, $n=5$ prevents premature alignment with immature features, and $M=4$ ensures sufficient interaction with high-level representations, and $\mu_1=0.1, \mu_2=1$ prove optimal for balancing the training objectives.

\begin{figure*}[t]
    \centering
    \includegraphics[width=\textwidth]{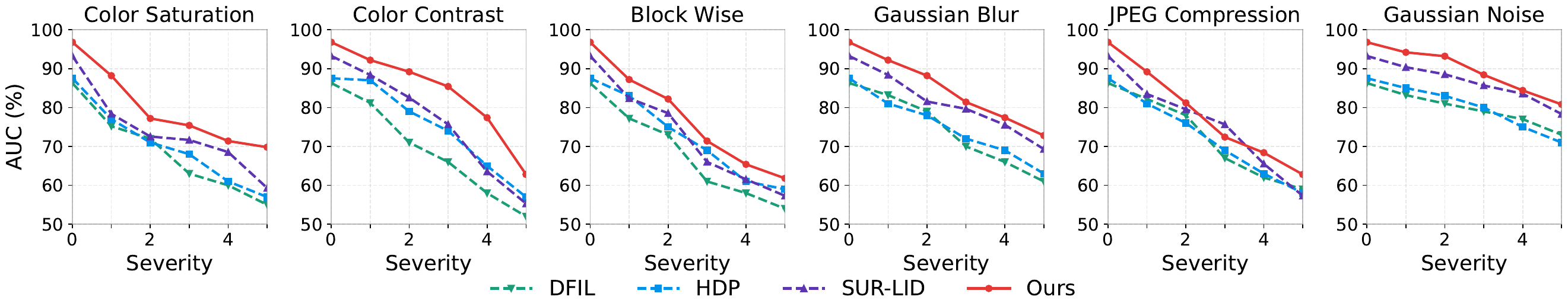}
    \caption{Robustness under unseen perturbations (following Protocol 1, average AUC is used as the evaluation metric).}
    \vspace{-5pt}
    \label{fig:robust}
\end{figure*}

\begin{table}[H]
\centering
\small
\caption{Ablation study (AUC) for each proposed component. \textbf{Bold} indicates the best performance.}
\renewcommand{\arraystretch}{1} 
\newcolumntype{Y}{>{\centering\arraybackslash}X} 
\begin{tabularx}{\linewidth}{l Y Y Y Y Y}
\toprule
Variant & SDv21 & FF++ & DFDCP & CDF & Avg. \\
\midrule
w/o All            & 0.9671 & 0.8746 & 0.9035 & 0.9513 & 0.9241 \\
w/o ADH            & 0.9897 & 0.9161 & 0.9280 & 0.9546 & 0.9471 \\
w/o APA            & 0.9896 & 0.9156 & 0.9283 & 0.9597 & 0.9483 \\
w/o $\mathcal{L}_{ind}$ & 0.9897 & 0.9173 & 0.9316 & 0.9706 & 0.9523 \\
\midrule
\textbf{Ours}      & \textbf{0.9988} & \textbf{0.9327} & \textbf{0.9551} & \textbf{0.9879} & \textbf{0.9686} \\
\bottomrule
\end{tabularx}
\label{tab:ablation}
\end{table}

\begin{table}[H]
\centering
\small
\caption{Performance comparisons (AUC) of APA injection layers. Bold indicates the best performance.}
\renewcommand{\arraystretch}{1} 
\newcolumntype{Y}{>{\centering\arraybackslash}X} 
\begin{tabularx}{\linewidth}{l Y Y Y Y Y}
\toprule
Depth & SDv21 & FF++ & DFDCP & CDF & Avg. \\
\midrule
Low            & 0.9108 & 0.7003 & 0.8751 & 0.9695 & 0.8639 \\
Medium         & 0.9833 & 0.9092 & 0.9258 & 0.9893 & 0.9519 \\
Multi          & 0.9466 & 0.8144 & 0.9011 & 0.9639 & 0.9065 \\
All            & 0.9597 & 0.9073 & 0.9236 & 0.9726 & 0.9408 \\
\midrule
High (Ours)      & \textbf{0.9988} & \textbf{0.9327} & \textbf{0.9551} & \textbf{0.9879} & \textbf{0.9686} \\
\bottomrule
\end{tabularx}
\label{tab:apa}
\end{table}

\vspace{-0.2cm}

\begin{figure}[H]
    \centering
    \small
    \begin{subfigure}{0.48\linewidth}
        \centering
        \includegraphics[width=\linewidth]{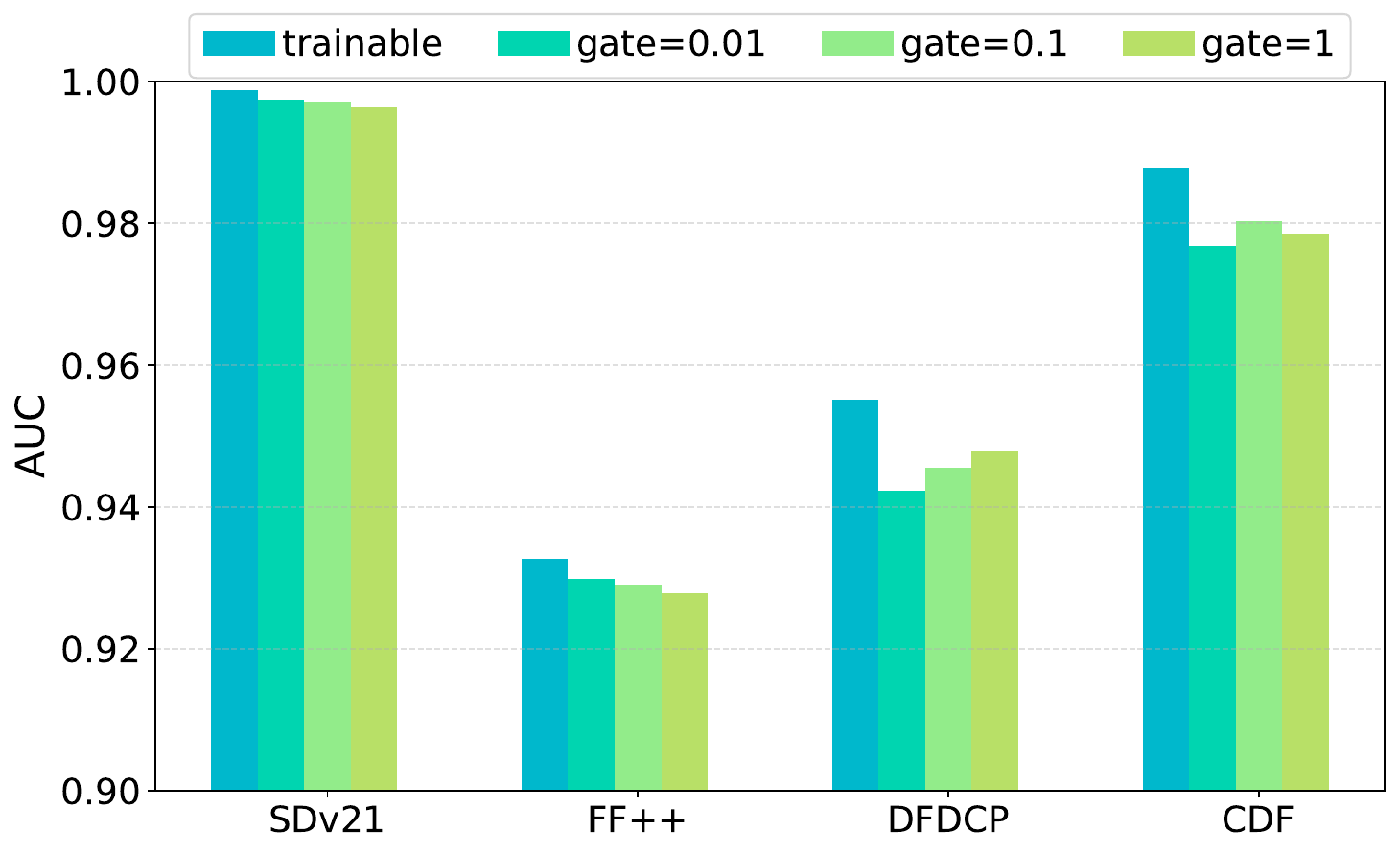}
        \caption{Protocol~1}
    \end{subfigure}
    \hfill
    \begin{subfigure}{0.48\linewidth}
        \centering
        \includegraphics[width=\linewidth]{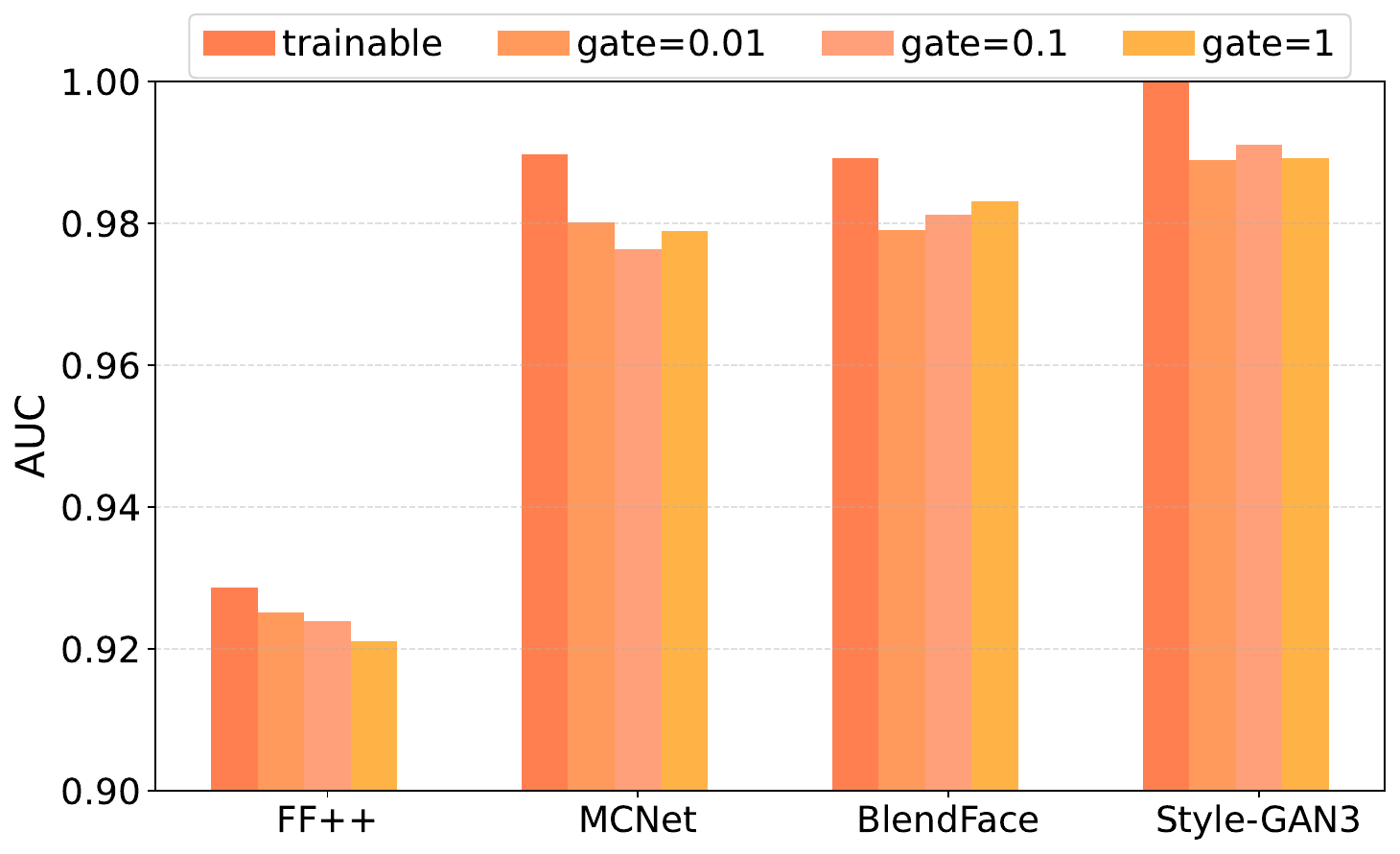}
        \caption{Protocol~1}
    \end{subfigure}
    \vspace{-0.1cm}
    \caption{Ablation study on the gating strategy within the Artifact-Probe Attention (APA) module.}
    \label{fig:gate}
\end{figure}

\noindent \textbf{Effect of APA Gating Strategy.}
We evaluate the impact of the semantic injection gate within the APA module by comparing fixed scales $\{0.01, 0.1, 1\}$ against a learnable parameter.
Unlike static settings that impose a rigid injection intensity, as shown in Fig.~\ref{fig:gate}, the learnable strategy yields superior performance, outperforming fixed scalars that suffer from either insufficient guidance (gate = $0.01$) or excessive feature perturbation (gate = $1$).
This confirms that adaptively modulating the injection ratio enables the model to dynamically balance semantic anchor integration with visual preservation, ensuring optimal fine-grained alignment without disrupting the underlying feature topology.

\begin{table}[H]
\centering
\small
\caption{Performance (AUC) under different classification-head alignment method. Bold indicates the best result.}
\renewcommand{\arraystretch}{1} 
\newcolumntype{Y}{>{\centering\arraybackslash}X} 
\begin{tabularx}{\linewidth}{l Y Y Y Y Y}
\toprule
Method & SDv21 & FF++ & DFDCP & CDF & Avg. \\
\midrule
LERP          & 0.9905 & 0.9314 & 0.9421 & 0.9792 & 0.9608 \\
EMA           & 0.9902 & 0.9286 & 0.9403 & 0.9807 & 0.9599 \\
WM            & 0.9935 & 0.9291 & 0.9462 & 0.9774 & 0.9615 \\
\midrule
\textbf{Ours}  & \textbf{0.9988} & \textbf{0.9327} & \textbf{0.9879} & \textbf{0.9686} & \textbf{0.9686} \\
\bottomrule
\end{tabularx}
\label{tab:adh}
\end{table}

\vspace{-0.3cm}

\begin{table}[H]
\centering
\caption{Performance (Avg. AUC on Protocol~1) under different parameter settings. \textbf{Bold} indicates the best result.}
\small
\renewcommand{\arraystretch}{1}
\newcolumntype{Y}{>{\centering\arraybackslash}X} 
\begin{tabularx}{\linewidth}{c | Y Y Y Y Y}
\toprule
\textbf{Parameter} & \multicolumn{5}{c}{\textbf{Settings \& Performance}} \\
\midrule
\multirow{2}{*}{$n$} 
& 0 & 5 & 10 & 15 & 20 \\
\cmidrule(lr){2-6}
& 0.9582 & \textbf{0.9686} & 0.9613 & 0.9601 & 0.9593 \\
\midrule
\multirow{2}{*}{$N$} 
& 1 & 2 & 3 & 5 & 10 \\
\cmidrule(lr){2-6}
& 0.9588 & 0.9621 & \textbf{0.9686} & 0.9614 & 0.9522 \\
\midrule
\multirow{2}{*}{$M$} 
& 1 & 2 & 3 & 4 & 5 \\
\cmidrule(lr){2-6}
& 0.9603 & 0.9611 & 0.9632 & \textbf{0.9686} & 0.9651 \\
\midrule
\multirow{2}{*}{$\mu_1$} 
& 0.01 & 0.05 & 0.1 & 0.2 & 0.5 \\
\cmidrule(lr){2-6}
& 0.9592 & 0.9621 & \textbf{0.9686} & 0.9576 & 0.9531 \\
\midrule
\multirow{2}{*}{$\mu_2$} 
& 0.1 & 0.5 & 1 & 1.5 & 2 \\
\cmidrule(lr){2-6}
& 0.9498 & 0.9572 & \textbf{0.9686} & 0.9581 & 0.9486 \\
\bottomrule
\end{tabularx}
\label{tab:params}
\end{table}

\subsection{Generalization and Robustness Evaluations}

To ensure a fair comparison, we standardize the backbone for all baseline methods to ViT-L/14, eliminating performance disparities caused by different backbones.

\noindent \textbf{Cross-Dataset Generalization.}
To rigorously evaluate the generalization capability of AIFIND against unseen domains, we conduct cross-dataset experiments. 
The model, fully trained under Protocol 1, is directly evaluated on four unseen benchmarks: DeepFakeDetection (DFD)~\cite{dfd}, UniFace~\cite{uniface} (from DF40~\cite{df40}), SDv15~\cite{sdv21} (from DiffusionFace~\cite{diffusionface}), and FakeAVCeleb (FAVC)~\cite{fakeavceleb}.
These datasets encompass a wide spectrum of generative mechanisms, ranging from conventional face swapping and GAN-based synthesis to diffusion-based text-to-image generation, which imposes significant challenges, requiring the model to overcome substantial domain shifts and rely on intrinsic, transferable forensic features rather than dataset-specific artifacts.

\begin{figure*}[t]
    \centering
    \small
    \begin{subfigure}{0.24\textwidth}
        \centering
        \includegraphics[width=\linewidth]{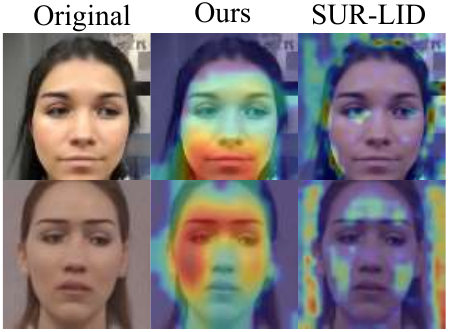}
        \caption{SDv2.1}
    \end{subfigure}
    \hfill
    \begin{subfigure}{0.24\textwidth}
        \centering
        \includegraphics[width=\linewidth]{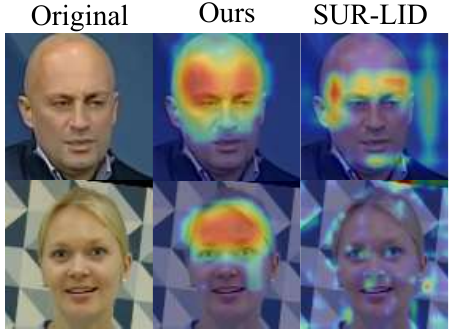}
        \caption{FF++}
    \end{subfigure}
    \hfill
    \begin{subfigure}{0.24\textwidth}
        \centering
        \includegraphics[width=\linewidth]{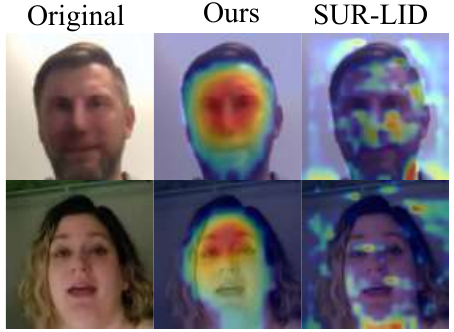}
        \caption{DFDCP}
        \label{fig:gradcam_c}
    \end{subfigure}
    \hfill
    \begin{subfigure}{0.24\textwidth}
        \centering
        \includegraphics[width=\linewidth]{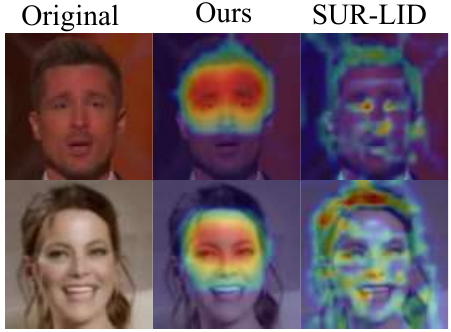}
        \caption{CDF}
        \label{fig:gradcam_d}
    \end{subfigure}

    \caption{Visualization of Grad-CAM heatmaps across different datasets.}
    \label{fig:gradcam}
\end{figure*}

\begin{figure*}[t]
    \centering

    \hspace{-12mm}
    \begin{subfigure}[b]{0.24\textwidth}
        \centering
        \includegraphics[height=3.2cm]{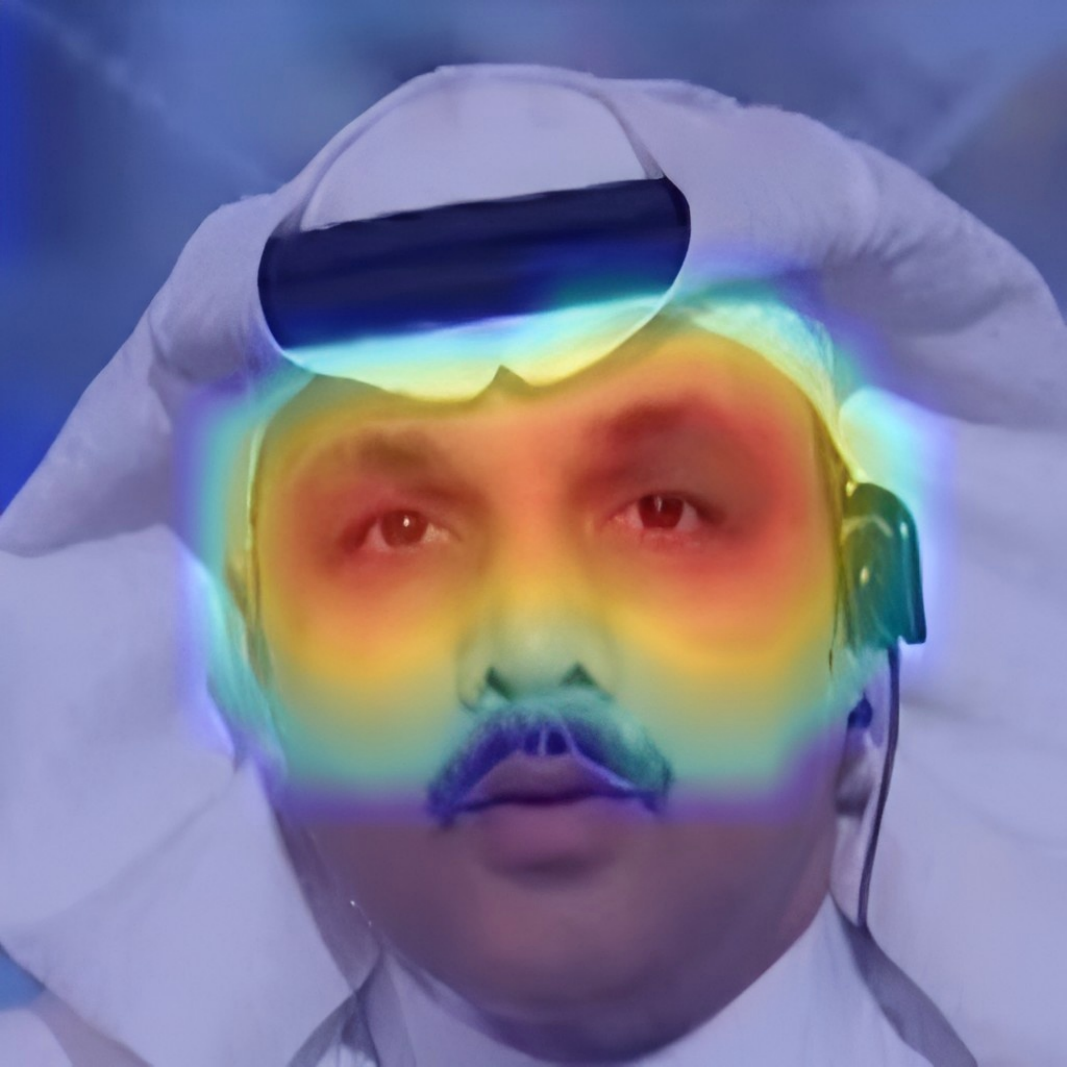}
        \caption{Grad-CAM Heatmap}
    \end{subfigure}\hspace{-7mm}
    \begin{subfigure}[b]{0.24\textwidth}
        \centering
        \includegraphics[height=3.2cm]{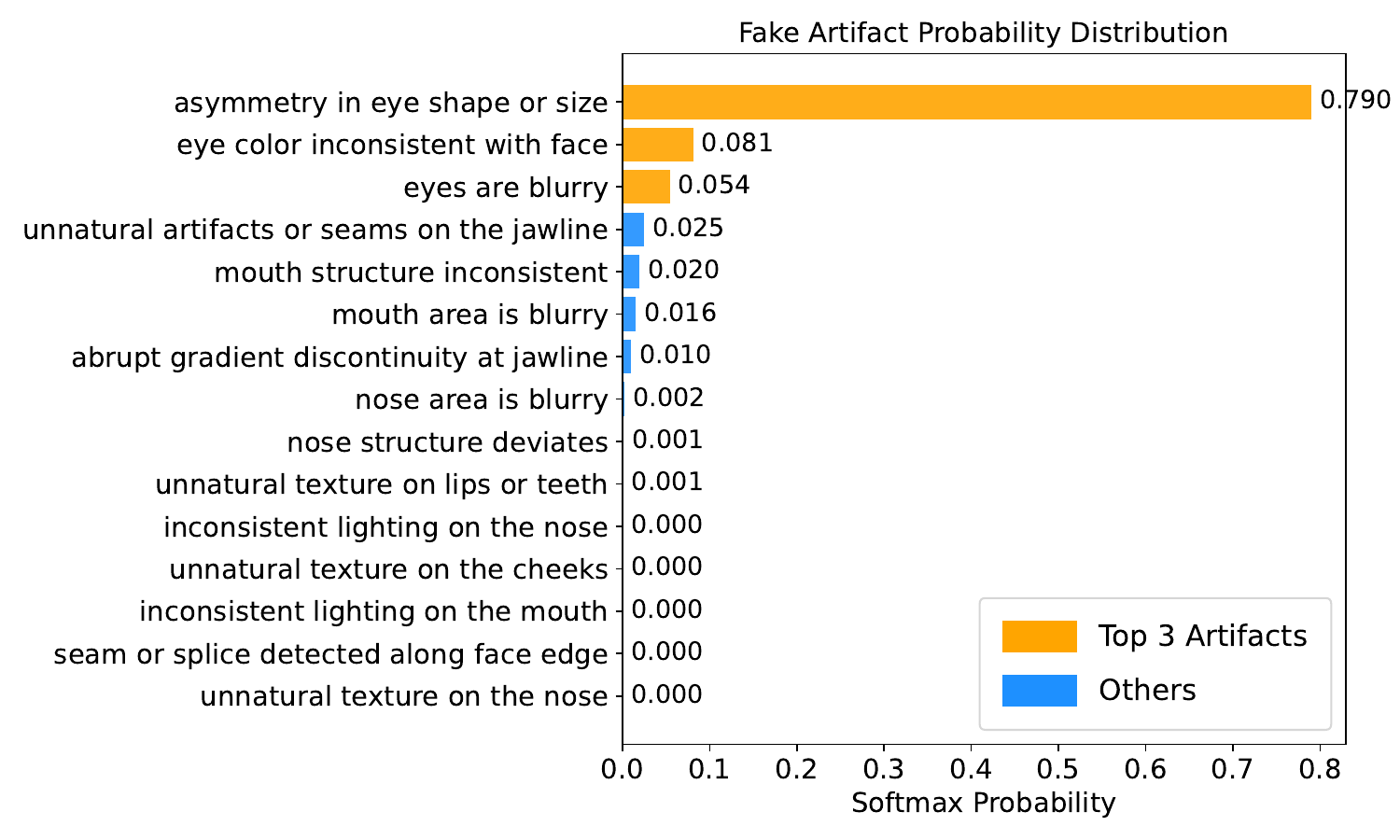}
        \caption{Probability Distribution}
    \end{subfigure}\hspace{8mm}
    \begin{subfigure}[b]{0.24\textwidth}
        \centering
        \includegraphics[height=3.2cm]{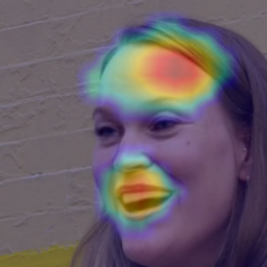}
        \caption{Grad-CAM Heatmap}
    \end{subfigure}\hspace{-7mm}
    \begin{subfigure}[b]{0.24\textwidth}
        \centering
        \includegraphics[height=3.2cm]{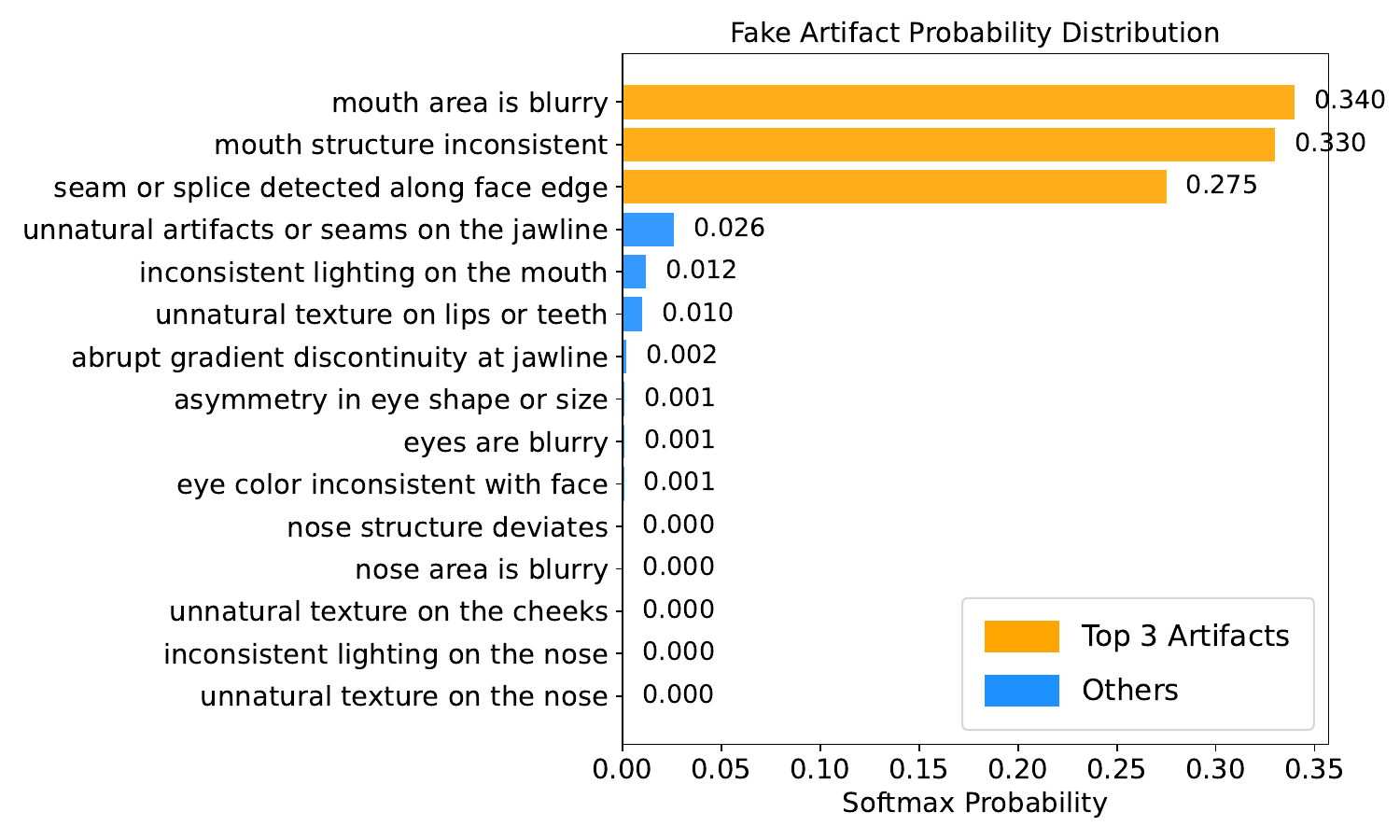}
        \caption{Probability Distribution}
    \end{subfigure}

    \caption{Consistency Between Attention Heatmaps and Fake Artifact Probability Distributions.}
    \label{fig:visual}
\end{figure*}

\begin{table}[H]
\centering
\small
\caption{Cross-dataset generalization results (AUC) on unseen datasets. \textbf{Bold} indicates the best performance.}
\renewcommand{\arraystretch}{1} 
\newcolumntype{Y}{>{\centering\arraybackslash}X} 
\begin{tabularx}{\linewidth}{l Y Y Y Y Y}
\toprule
Method & DFD & UniFace & SDv15 & FAVC & Avg. \\
\midrule
DFIL            & 0.8212 & 0.6549 & 0.8236 & 0.7157 & 0.7564 \\
HDP             & 0.8342 & 0.6977 & 0.8231 & 0.7495 & 0.7761 \\
SUR-LID            & 0.8825 & 0.8279 & 0.8522 & 0.8143 & 0.8442 \\
\midrule
\textbf{Ours}      & \textbf{0.9332} & \textbf{0.8987} & \textbf{0.8847} & \textbf{0.8755} & \textbf{0.8980} \\
\bottomrule
\end{tabularx}
\label{tab:cross_generalization}
\end{table}

As shown in Tab.\ref{tab:cross_generalization}, our framework achieves superior cross-dataset generalization across all unseen benchmarks.
Unlike baselines that may overfit to dataset-specific patterns, AIFIND learns intrinsic and transferable artifact representations, enabling robust detection even in unknown domains.

\noindent \textbf{Robustness Evaluations}
For robustness evaluation, we adopt the rigorous perturbation protocols from~\cite{deeperforensics}, which include five severity levels across six different perturbation types.
As shown in  Fig.\ref{fig:robust}, our model consistently achieves higher AUC scores across all perturbation levels compared to other baselines.
These results suggest that by aligning visual features with stable semantic anchors, AIFIND preserves discriminative ability under significant image distortions, ensuring reliable performance in practical scenarios.

\subsection{Visualizations}
We employ Grad-CAM~\cite{gradcam} to visualize the attention maps of the model trained under Protocol~1. 
As illustrated in Fig.~\ref{fig:gradcam}, the baseline SUR-LID~\cite{cheng2025stacking} exhibits relatively dispersed attention, often appearing distracted by the background or irrelevant facial areas. 
In contrast, our method tends to concentrate more on semantically sensitive facial components, such as the eyes and mouth, which are notoriously prone to manipulation artifacts. 
This improved focus suggests that the semantic anchors effectively guide the model to attend to critical forensic regions rather than low-level noise, validating that our linguistic supervision successfully directs visual attention to physically meaningful areas.

Furthermore, the results in Fig.~\ref{fig:visual} demonstrate a notable **spatial-semantic consistency** between the attention heatmaps and the inference outcomes. 
Specifically, the regions receiving high visual activation generally align with the artifact categories that yield elevated predicted probabilities. For instance, when the heatmap highlights the eye region, the probability score for eye-related artifact classes rises distinctively.
These observations indicate that our model learns to associate discriminative forensic traces with their corresponding semantic priors, thereby mitigating the risk of overfitting to spurious cues and enhancing the interpretability and reliability of the decision-making process.

\vspace{-0.5cm}

\section{Conclusion}
In this paper, we propose AIFIND, an Artifact-Aware Interpreting Fine-Grained Alignment framework. 
Unlike traditional methods that rely on sample replay, AIFIND leverages a semantic anchor library to guide the model in learning invariant forgery representations. 
Our method ensures that visual features are aligned with stable semantic anchors, effectively mitigating catastrophic forgetting. 
Extensive experiments demonstrate that our method achieves state-of-the-art performance and exhibits spatial-semantic consistency in visualization.
In the future, we plan to extend our framework to broader multimodal scenarios, exploring more adaptive semantic anchors via Large Vision-Language Models to tackle increasingly diverse forgery patterns in open-world settings.

\begin{acks}
This work is partially supported by the National Natural Science Foundation of China under Grants 62441232, 62476068, 62306092, 62502115, and projects ZR2025ZD01, ZR2024QF066, ZR2025QC1516 supported by Shandong Provincial Natural Science Foundation, and projects 2024DXZD0004 supported by Inner Mongolia Department of Science and Technology.
\end{acks}


\bibliographystyle{ACM-Reference-Format}
\bibliography{reference}

\end{document}